\documentclass[lettersize,journal]{IEEEtran}
\usepackage{cite}
\usepackage{amsmath,amssymb,amsfonts}
\usepackage{algorithmic}
\usepackage{graphicx}
\usepackage{textcomp}
\def\BibTeX{{\rm B\kern-.05em{\sc i\kern-.025em b}\kern-.08em
    T\kern-.1667em\lower.7ex\hbox{E}\kern-.125emX}}

\usepackage{xtab}
\usepackage{graphicx}
\usepackage{array} 
\usepackage{multirow} 
\usepackage[table,xcdraw]{xcolor} 
\usepackage{placeins} 
\usepackage{booktabs}    
\usepackage{ragged2e}
\usepackage{makecell}    
\usepackage{array}   
\usepackage[protrusion=true,expansion=true]{microtype}
\pdfoutput=1
\usepackage{tabularx}
\usepackage{url}
\usepackage[breaklinks=true]{hyperref}

\usepackage{placeins}  
\usepackage{float}     
\usepackage{colortbl}
\definecolor{darkyellow2}{RGB}{247, 234, 223} 
\definecolor{salmon2}{RGB}{150, 112, 126}
\definecolor{Gray}{gray}{0.9}
\definecolor{darkyellow}{RGB}{219, 181, 181} 
\definecolor{salmon}{RGB}{237, 230, 200}
\definecolor{darksalmon}{RGB}{201, 193, 160}
\usepackage{multirow}
\definecolor{darkgreen}{RGB}{180, 230, 198}
\definecolor{lightgreen}{RGB}{209, 233, 201}

\definecolor{lightred}{RGB}{255,211,222}
\definecolor{darkred}{RGB}{252,150,167}
 \definecolor{LightBlue}{rgb}{0.8,0.89,1} 
\definecolor{DarkBlue}{rgb}{0.57,0.71,0.82} 
\definecolor{MagLight}{rgb}{1, 0.89, 0.8}
\usepackage{enumitem}
  \makeatletter
\newcommand\notsotiny{\@setfontsize\notsotiny\@vipt\@viipt}
\makeatother
\usepackage[most]{tcolorbox}

\begin{document}
\title{\vspace{-5.5mm}Hierarchical Federated Foundation Models over Wireless Networks for Multi-Modal Multi-Task Intelligence: Integration of Edge Learning with D2D/P2P-Enabled Fog Learning Architectures     \vspace{-1mm}}

\author{Payam Abdisarabshali,~\IEEEmembership{Student Member,~IEEE,} Fardis Nadimi,~\IEEEmembership{Student Member,~IEEE,} Kasra Borazjani,~\IEEEmembership{Student \\ Member,~IEEE,} Naji Khosravan,~\IEEEmembership{Member,~IEEE,} Minghui  Liwang,~\IEEEmembership{Senior Member,~IEEE,} Wei Ni,~\IEEEmembership{Fellow,~IEEE,} Dusit Niyato,~\IEEEmembership{Fellow,~IEEE,} Michael Langberg,~\IEEEmembership{Fellow,~IEEE,} and Seyyedali Hosseinalipour,~\IEEEmembership{Senior Member,~IEEE}
        \vspace{-4.5mm}
}



\maketitle

\begin{abstract}
The rise of foundation models (FMs) has reshaped the landscape of machine learning. As these models continued to grow, leveraging geo-distributed data from wireless devices has become increasingly critical, giving rise to federated foundation models (FFMs).
More recently, FMs have evolved into multi-modal multi-task (M3T) FMs (e.g., GPT-4) capable of processing diverse modalities across multiple tasks, which motivates a new underexplored paradigm: M3T FFMs.
In this paper, we unveil an unexplored variation of M3T FFMs by proposing \textit{hierarchical federated foundation models ({\tt HF-FMs})}, which in turn expose two overlooked heterogeneity dimensions to fog/edge networks that have a direct impact on these emerging models: 
(i) heterogeneity in collected modalities and (ii) heterogeneity in executed tasks across fog/edge nodes.
 {\tt HF-FMs} strategically align the modular structure of M3T FMs, comprising modality encoders, prompts, mixture-of-experts (MoEs), adapters, and task heads, with the hierarchical nature of fog/edge infrastructures. Moreover, {\tt HF-FMs} enable the optional usage of device-to-device (D2D) communications, enabling horizontal module relaying and localized cooperative training among nodes when feasible. 
Through delving into the architectural design of {\tt HF-FMs}, we highlight their unique  capabilities along with a series of tailored future research directions. Finally, to demonstrate their potential, we prototype {\tt HF-FMs} in a wireless network setting and release the open-source code for the development of {\tt HF-FMs}  with the goal of fostering exploration in this untapped field (GitHub: {\footnotesize \url{https://github.com/payamsiabd/M3T-FFM}}).

\end{abstract}

\maketitle

\section{Introduction}
From autonomous vehicles to industrial sensors and wearables, modern wireless devices generate streams of high-dimensional multi-modal data (e.g., LiDAR, audio, video, telemetry), where integrating this geo-distributed data to train machine learning (ML) models is becoming central to enabling autonomy and real-time decision-making across wireless networks.
Nevertheless, conventional ML techniques require transmitting the collected raw data from wireless devices to centralized servers for model training, which is often infeasible due to latency bottlenecks and privacy risks. 

\begin{figure}
    \centering
    \includegraphics[width=\linewidth]{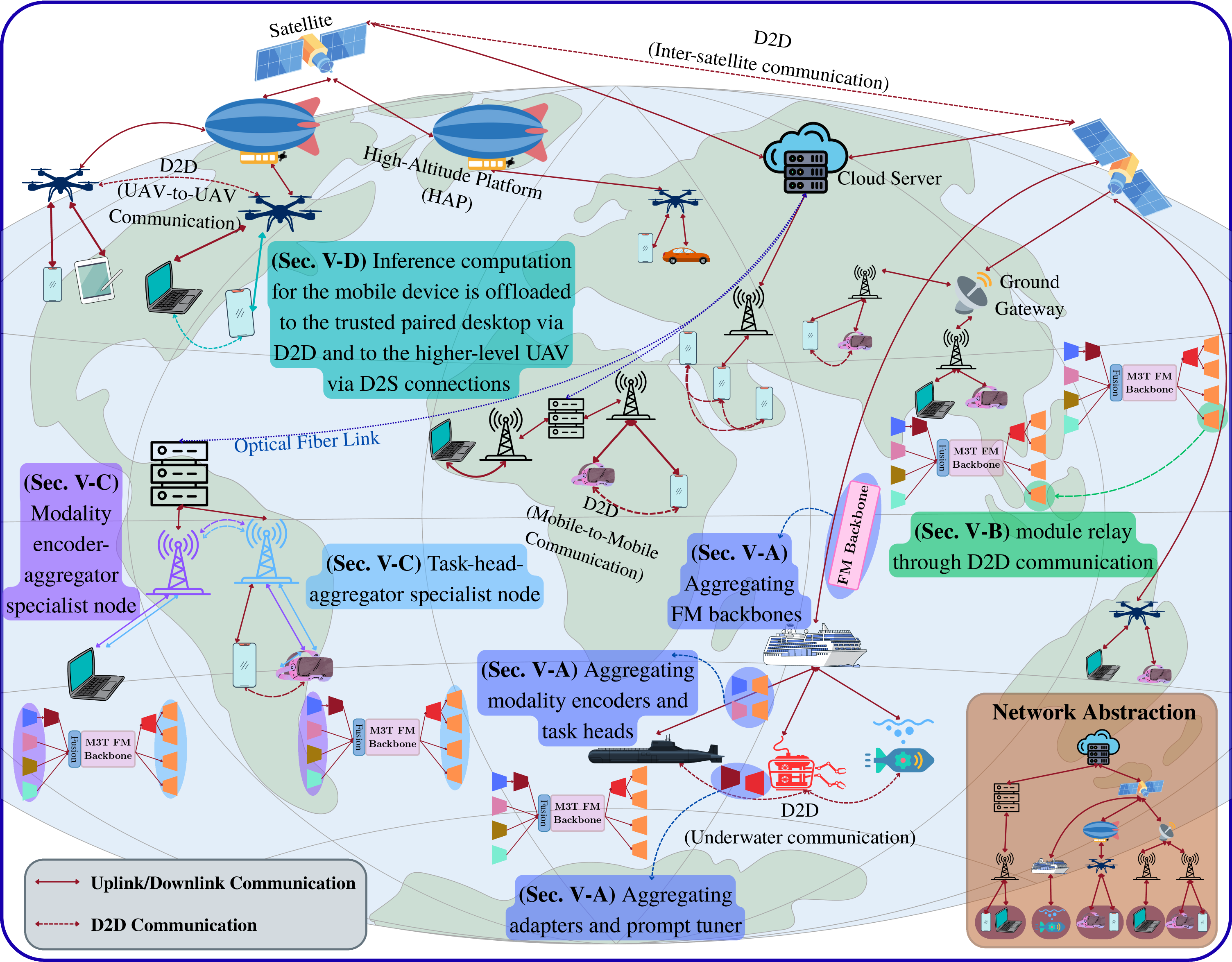}
    \vspace{-7mm}
    \caption{A schematic of a hierarchical/multi-tier fog network alongside its abstract representation (bottom right). The research directions discussed in Sec.~\ref{sec:challenges} are color-coded and highlighted in the illustration.}
    \label{fig:challenges}
    \vspace{-5mm}
\end{figure}

To overcome these barriers, the \textit{edge learning} paradigm has emerged, where ML models are trained locally on edge devices, typically via federated learning (FL)~\cite{tak2020federated}. While edge learning mitigates data transfer overhead and privacy risks, its reliance on a \textit{star-shaped communication topology}, where devices communicate directly with a central server, can limit its scalability.
To address this, the concept of \textit{fog learning}, introduced in~\cite{hosseinalipour2021federated}, extends edge learning by embracing a topology-aware, multi-layered approach to distributed ML. Fog learning enables a hierarchical architecture that includes various fog nodes such as base stations, access points, edge servers, drones, high-altitude platforms, and routers that act as intermediate model aggregators and/or relay points. As illustrated in Fig.~\ref{fig:challenges}, fog nodes collaboratively form a \textit{multi-layer/tier network}, facilitating layer-wise model relaying and aggregation. Moreover, when equipped with relevant local data, fog nodes can actively participate in model training,  enhancing the system-wide learning scalability and efficiency.
Fog learning further exploits both \textit{vertical} parameter aggregation across network layers and \textit{horizontal} collaboration between wireless/wired nodes located in the same network layer via device-to-device (D2D) or peer-to-peer (P2P) communications, aiming to reduce the usage of licensed spectrum, accelerate training convergence, and support adaptive distributed ML under dynamic mobility and network resource conditions \cite{hosseinalipour2022multi}.


Both edge and fog learning were initially designed for narrow, task-specific ML models, reflecting the state of earlier-generation AI systems. Nevertheless, the advent of \textit{foundation models (FMs)} such as GPT-3 has marked a significant paradigm shift. These models, often comprising billions/trillions of parameters, are pre-trained on massive and diverse datasets using self-supervised (contrastive learning) objectives, offering broad generalization across a wide array of applications.
Although FMs were initially trained in \textit{centralized} settings using web-scale datasets and high-performance computing clusters, the \textit{geo-distributed} and \textit{privacy-sensitive} nature of user data, particularly in wireless networks, challenges the practicality of this centralized paradigm.
These constraints have spurred interest in adapting FMs to distributed ML frameworks. 
Specifically, FL approaches can enable fine-tuning and continual refinement of FMs: a capability that has led to the emergence of \textit{federated foundation models (FFMs)}~\cite{ren2025advances}. FFMs combine the generalizability of FMs with the distributed, privacy-preserving, and scalable training capabilities of FL.

Although several studies have investigated FFMs, the majority have centered on FL-driven training/fine-tuning of large language models (LLMs), which are FMs that operate primarily over textual inputs and language-based tasks. This emphasis is largely due to the early success of LLMs as the prototypical FMs. In this work, we adopt a more forward-looking perspective by focusing on the latest evolution of FMs: \textit{multi-modal multi-task (M3T) FMs}, exemplified by architectures such as Perceiver IO, Flamingo, and GPT-4\cite{li2024multimodal}. These models are capable of jointly processing diverse data modalities (e.g., video, audio, and text) and supporting a wide range of tasks (e.g., text generation, image classification, and image synthesis), thereby enabling emerging applications at the wireless edge, including embodied AI, autonomous vehicles, and extended reality systems~\cite{nadimi2025multi}.

Unlike traditional narrow models, M3T FMs are inherently \textit{modular} and \textit{compositional}, with architectures explicitly designed to support module-level training and fine-tuning. This architectural flexibility raises a natural and timely research question: \textit{how can the hierarchical topology of fog/edge networks, and possible D2D local cooperation among nodes, be harnessed and adapt to the modular design of M3T FMs?}
To address this question, we propose \textit{hierarchical federated foundation models {\tt (HF-FMs)}}: a new learning paradigm that broadens the reach of FFMs by tightly integrating the modular structure of M3T FMs with the layered architecture of fog/edge infrastructures. {\tt HF-FMs} further extend FFMs' capabilities by enabling optional D2D cooperation among fog/edge nodes during model training/fine-tuning and aggregation. In essence, {\tt HF-FMs} support distributed, scalable, and context-aware learning across fog/edge hierarchies, wherein each network layer contributes to model refinement based on its available resources, target ML tasks, communication and computation capabilities, and access to modality-specific data. 


Subsequently, this work is positioned as a \textit{vision paper}, aiming to unveil the concept of {\tt HF-FMs}, demonstrate their potential to reshape distributed AI systems, and spark future research in this largely unexplored domain. Our key contributions are summarized as follows:

\begin{itemize}[leftmargin=3.9mm]
    \item \textit{Introduction of {\tt HF-FMs}:} We introduce {\tt HF-FMs}, novel architectural designs that tightly integrate the modular training/fine-tuning procedures of M3T FFMs with hierarchical fog/edge computing infrastructures. {\tt HF-FMs} further allow for the use of D2D cooperation between nodes during the training/fine-tuning processes of M3T FFMs.
    
    \item \textit{Module-Wise Circulation:} We present a decomposition of {\tt HF-FM} procedures, illustrating how modules such as prompts, modality-specific encoders, task heads, mixture-of-experts (MoEs), and adapters can be distributed, trained, and aggregated across multi-tier fog/edge network hierarchies.
    
    \item \textit{Emerging System Capabilities:} We introduce four emerging system capabilities enabled by {\tt HF-FMs}: (i) asymmetric aggregation, (ii) module relaying, (iii) node specialization, and (iv) collaborative inference. For each, we articulate associated opportunities and open research directions.
    
    \item \textit{Benchmarking Framework:} We prototype {\tt HF-FMs} within a wireless network setting and release the open source code for the development of {\tt HF-FM}, with the goal of catalyzing research and exploration in this nascent area.
\end{itemize}

In the rest of this paper, due to the structural overlap between fog and edge networks, where fog networks subsume edge network elements, we will use the generic reference to ``fog/edge" nodes unless an explicit distinction is needed.

\section{Background and Related Work}
While {\tt HF-FMs} represent a novel framework, they have innate roots in existing methodologies. Below, we delineate the key differences between {\tt HF-FMs} and these prior approaches.

\subsubsection{Conventional FL and Hierarchical FL (HFL)}
Traditionally, FL operates under a \textit{device-to-server ``star" topology}, where devices train \textit{local models} and periodically transmit model/gradient parameters to a server for aggregation~\cite{tak2020federated}. The server combines these updates (e.g., via weighted averaging) to produce a \textit{global model} and broadcast it to all devices for the next round of training. This process continues until a convergence criterion is satisfied. However, such a centralized model/gradient aggregation often becomes infeasible in large-scale networks as (i) it can impose severe traffic on network backhaul and (ii) some devices may lack reliable access to a central server. 
Subsequently, HFL emerged as an extension of FL, introducing multi-level aggregation across a device–fog/edge–cloud hierarchy~\cite{liu2020client,wahab2021federated}. In HFL, devices send their local model updates to nearby fog/edge nodes (e.g., edge servers, access points, base stations, or drones) for frequent, localized aggregation. 
Aggregated models are then selectively forwarded to higher-tier nodes for broader aggregations (e.g., metro-core or cloud servers) at lower frequencies.
Fog learning builds upon the HFL paradigm by incorporating D2D communications into the learning architecture. This enables low-power cooperative model exchanges between proximate devices/nodes~\cite{hosseinalipour2022multi}.

Despite the promising advances of HFL and fog learning for narrow models, extending these paradigms to the M3T FFM setting, where models are large-scale, modular, and compositional, remains open. In this work, we bridge this gap by introducing \textit{{\tt HF-FMs}}, designed to integrate the model complexity and modular training flexibility of M3T FMs into the hierarchical fog/edge computing infrastructure. While optional, the {\tt HF-FM} framework enables the use of D2D communications when nodes are able to operate in D2D mode, offering additional latency and energy savings.

\subsubsection{FMs, FFMs, and their M3T Variants}
The evolution of FMs began with LLMs, such as GPT-3, which demonstrated remarkable generalization across a broad range of natural language processing tasks. However, LLMs were inherently text-centric and lacked native multi-modal support.
Recently, the development of M3T FMs, such as Flamingo, Perceiver IO, and GPT-4 \cite{li2024multimodal}, has marked a significant paradigm shift. These models are architected to simultaneously process multiple data modalities (e.g., text, images, audio, and video) and support diverse ML tasks (e.g., text generation and image classification).
Despite their architectural diversity, nearly all variants of FMs rely on a \textit{centralized training/fine-tuning} strategy, leveraging either large-scale curated datasets or web-scraped data.

Recently, the urge to utilize geo-distributed data, as seen in applications such as autonomous driving, has motivated a growing body of research into FL-based training/fine-tuning of FMs, giving birth to FFMs. Early efforts have explored various techniques to reduce the computational and communication overhead of training large models in distributed settings. These include parameter-efficient fine-tuning methods, such as low-rank adaptation (LoRA), prompt tuning, and instruction tuning (see \cite{ren2025advances} and references therein).
However, the majority of existing FFM frameworks adopt a vanilla device-to-server \textit{star aggregation topology} and focus on LLMs, overlooking both the hierarchical structure of real-world fog/edge networks and the opportunities for cooperative D2D communications among fog/edge elements. Moreover, the modular and compositional nature of M3T FMs can be synergistic with hierarchical network architectures, which is yet to be studied. This work addresses these gaps by exploring the adaptation of  M3T FFMs to D2D-enabled hierarchical fog/edge networks through {\tt HF-FMs}.



\begin{figure}[t]
    \centering
    \includegraphics[width=\linewidth]{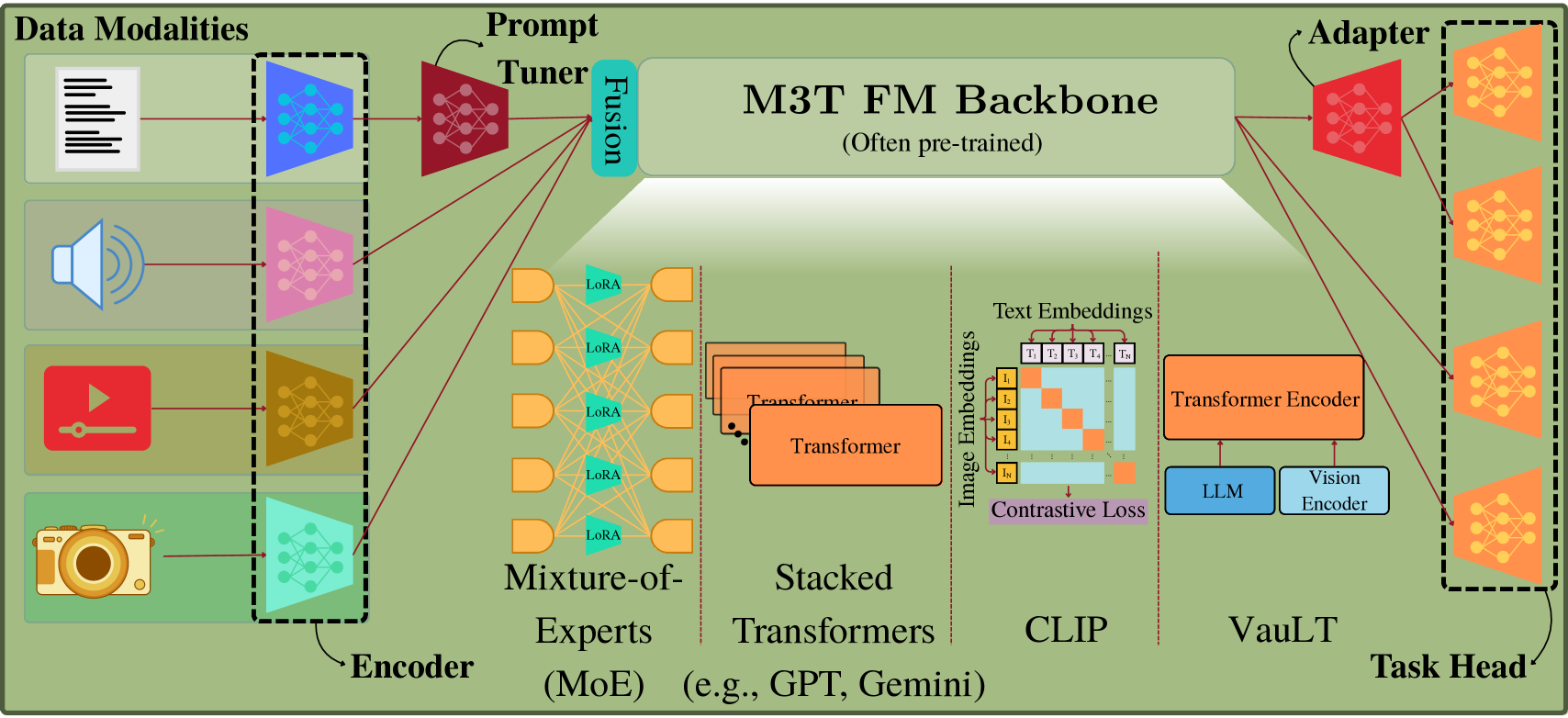}
    \vspace{-7mm}
    \caption{A schematic of M3T FM architecture alongside its backbone variants.}
    \label{fig:fm-model}
        \vspace{-3mm}
\end{figure}
\section{Overarching Learning Architecture}
To ground our discussions, we begin by examining the architecture of recent M3T FMs, followed by a brief review of device orchestration frameworks used in FFM literature. 

\subsubsection{M3T FMs Structure}
M3T FMs represent a broad and evolving family of architectures under active development in both academia and industry. To facilitate our exposition, we abstract a representative set of M3T FM architectures (see Fig.~\ref{fig:fm-model}) that reflect current design trends\cite{chen2024disentanglement,brown2020language,radford2021learning,xiao2024configurable,chen2024feddat}, and detail their constituent modules below.

\textbf{1. Modality-Specific Encoders:} 
M3T FMs typically initiate data processing through dedicated \textit{modality encoders}, transforming raw sensory signals (e.g., text, images, and audio) into \textit{embeddings}, which are then input/fed to the backbone.

\textbf{2. Shared Backbone:} A shared backbone is responsible for multi-modal fusion and contextual learning from the input embeddings. This backbone is the cornerstone of M3T FMs, enabling the model to learn contextual meanings that can be generalized across diverse tasks and users. Representative architectures of M3T FMs' backbones are shown in  Fig.~\ref{fig:fm-model}, and detailed below:

\begin{itemize}[leftmargin=3mm]
    \item \textit{Transformer-Based Backbones (e.g., GPT, Gemini, and VauLT)~\cite{brown2020language,chen2024feddat}:} A unified stack of transformer layers learns from and processes input embeddings conjunctly.
    
    \item \textit{Dual or Multi-Encoder Fusion (e.g., CLIP)~\cite{radford2021learning}:} Each modality is processed through its own stream (e.g., a transformer), and the results are further processed for contextual learning, often using contrastive learning techniques.
    
    \item \textit{Mixture-of-Experts (MoE)~\cite{chen2024disentanglement}:} A set of expert networks/modules (e.g., neural networks or transformers) process the input embeddings. MoEs enable selective activation of modality-specific or task-specific pathways within the model. In this architecture, upon inputting a datapoint, an internal gating network determines the most suitable subset of experts to engage/activate based on the extracted embeddings.
\end{itemize}

\textbf{3. Task-Specific Heads:} Located at the end of the model, \textit{task heads} are responsible for mapping the representations obtained from the backbone to specific ML task outputs (e.g., text generation or image classification). 

\subsubsection{Training \& Fine-Tuning}
Since training the full FM (or M3T FM) can be infeasible for devices due to the model's scale, devices often \textit{fine-tune} the models via four main techniques. \textit{(i) Adapters and LoRA modules:} These are compact \textit{trainable modules/parameters} inserted between layers of the model. \textit{(ii) Prompt-Tuning:} Prompts are \textit{learnable tokens} appended to the input or injected at key points within the model. By learning only these tokens, the model's output can be adapted to an unseen data distribution. \textit{(iii) Selected MoE training:} In resource-constrained environments, \textit{a small subset of the experts} (i.e., the most relevant ones) can be trained locally while freezing the remaining experts. \textit{(iv) Encoder and Task Head Training:} Upon experiencing a major shift in the distribution of a modality (task) or the emergence of a new modality (task) type, modality encoders (task heads) can be selectively trained. Subsequently, FFMs expand these techniques to the distributed learning setups: devices typically engage in local training/fine-tuning of their FMs (or M3T FMs) and send their training results (i.e., LoRA module/weights) to a server for model/module aggregations and broadcast, following the conventional FL protocol. It is worth mentioning that the literature on M3T FFMs is scarce~\cite{chen2024disentanglement,chen2024feddat}, as FFMs initially focused on distributed training/fine-tuning of LLMs~\cite{ren2025advances}.


\subsubsection{Interpretation of Modularity}
The compositional/modular structure of M3T FMs arises from the fact that individual modules (i.e., prompts, LoRA parameters, task heads, encoders, or expert networks) can be trained, updated, or personalized independently. This inherent modularity enables decomposable FFM operations, allowing nodes to train and exchange only the relevant modules, thereby reducing the model training overhead while enabling model personalization.

$\star$ We move beyond the traditional star-topology FL architecture employed in existing FFM (and M3T FFM) studies and explore how should {\tt HF-FMs} accommodate the above-described modularity in the topological architectures of fog/edge networks. 
Specifically, the modular structure of M3T FMs enables fine-grained controls over local training and fine-tuning: modality encoders can be selected based on the data modalities collected at fog/edge nodes, task heads can be aligned with the node’s downstream inference or control tasks, and tuning strategies (e.g., prompt tuning) can be customized to local constraints.
This flexibility, coupled with the inherent imbalance in both data modalities and tasks across nodes, \textit{introduces two underexplored dimensions of heterogeneity} that fundamentally impact the operation of M3T FFMs in fog/edge environments:
\textit{(i) modality heterogeneity,} stemming from the diverse sensing capabilities of fog/edge nodes, and
\textit{(ii) task heterogeneity,} driven by the varied functional roles (e.g., perception, control, and actuation) these nodes serve.
These dual heterogeneities challenge core assumptions of classical FL and HFL and calls for the design of nuanced operations in {\tt HF-FMs}, including module circulation across nodes, layer-aware training/fine-tuning schedules, and aggregation mechanisms tailored to the heterogeneity profile of the network, which are discussed in this work.

\textbf{Our Vision:}
The native heterogeneity of fog/edge network elements across \textit{communication, computation, and storage} (CCS) resources and the non-iid (non-independent and identically distributed) nature of their data, when complemented by the dual heterogeneity introduced by M3T FFMs (i.e., modality and task heterogeneities) poses two promising research directions: 
 these heterogeneities can be addressed via \textit{(i) dynamic network/device orchestration},  where the fog/edge infrastructure is dynamically adapted  to support them, and \textit{(ii) novel learning techniques} that enhance the core ML framework.
In this work, we focus on the former and leave the complementary direction of heterogeneity-aware ML algorithm designs to future work.

\textbf{Terminology:} Henceforth, we use ``FM" to denote ``M3T FMs", and ``FFM" to denote ``M3T FFMs", as our focus is exclusively on M3T-based architectures. All references to the ``model" refer to the underlying M3T FM architecture.


\section{{\tt HF-FMs}: Overview and Prototyping}
  In their \textit{simplest instantiations}, as depicted in Fig.~\ref{fig:challenges}, we showcase {\tt HF-FMs} by combining the hierarchical aggregation of HFL with the cooperative capabilities of fog learning. Such an extension establishes a mapping among: (i) \textit{modular training} of FFMs, (ii) the above-described \textit{task and modality heterogeneity}, (iii) the inherent \textit{heterogeneity of fog/edge nodes} in terms of CCS constraints, and (iv) the \textit{hierarchical topology} of fog/edge infrastructures. 
  Specifically, {\tt HF-FMs} can be realized by clustering fog/edge nodes based on spatial or network proximity, selecting a cluster head for each cluster, and performing module-specific training (e.g., adapters) at the cluster level. The resulting component updates can then be aggregated at the cluster head with high-frequency and propagated hierarchically to higher-tier nodes (e.g., regional/cloud servers) at lower frequencies for broader aggregations. Additionally, D2D communications can be leveraged to (i) facilitate multi-hop model relaying between  nodes and/or (ii) form cooperative decentralized model consensus (e.g., using distributed consensus~\cite{hosseinalipour2022multi}), reducing  reliance on resource-intensive inter-layer communications.  

However, whether such a hierarchical D2D-enabled realization leads to tangible performance gains for {\tt HF-FMs} remains an open question, due to the distinct dynamics of modular training (i.e., the use of fine-tuning techniques on specific modules compared to full-model training) and aggregation in FMs. Motivated by this, we present a case study below that empirically investigates the feasibility and potential benefits of the {\tt HF-FMs} framework.

\subsection{Case Study}
We consider a three-tier fog/edge network comprising $40$ edge devices (tier 1), $10$ edge servers (tier 2), and a single cloud server (tier 3). The edge devices are partitioned into $10$ clusters ($4$ nodes are considered in each cluster, forming a Random Geometric Graph~\cite{hosseinalipour2022multi}). We investigate {\tt HF-FMs} on two heterogeneous Visual Question Answering (VQA) datasets/tasks~\cite{chen2024feddat}: (i) Art and (ii) General Visual Question Answering (GQA).  The Art dataset focuses on VQA in the context of artwork, requiring nuanced reasoning about abstract, stylistic, and often subjective visual elements. In contrast, GQA involves real-world images and is designed to test compositional reasoning over object attributes, spatial relationships, and factual knowledge grounded in everyday scenes. 

\subsection{Simulation Setup}
The D2D network topology, uplink/downlink channel models, and corresponding data rate computations are implemented following~\cite{lin2021semi,liu2020client}. 
Each dataset/task is distributed across nodes in a non-iid manner (using Dirichlet distribution~\cite{wahab2021federated} with concentration parameter $0.1$). We adopt ViLT as the backbone of the FMs deployed/trained at nodes. Following the fine-tuning method developed in~\cite{chen2024feddat}, the model is augmented with lightweight \textit{adapters}, integrated into each transformer layer, and task-specific heads. Each adapter follows a bottleneck design with a hidden size of $256$: it includes a down-projection layer, a GELU activation, and an up-projection layer. To further enrich the model’s representational capacity, we employ Double-Adapter modules~\cite{chen2024feddat} that combine the outputs of two parallel adapters. 
The total size of the adapters and classification heads transmitted over the network is 6 MB, whereas the size of the ViLT backbone is 328 MB. 
Our implementations can be found at  GitHub: {\small \url{https://github.com/payamsiabd/M3T-FFM}}. 
We examine the following methods:

\textit{1) Conventional FFM over Star Topology:} Inspired by~\cite{chen2024feddat}, after $E_{\mathsf{Local}}=1$ epochs of local training, fine-tuned parameters are sent to the cloud and aggregated using \textit{FedAvg}.


\textit{2) {\tt \textit{HF-FMs}} without D2D Links:} After $E_{\mathsf{Local}}=1$ epoch of local training, the fine-tuned parameters are sent to the edge servers and aggregated using FedAvg. For cross-cluster knowledge sharing, after every $E_{\mathsf{Agg}}$ edge aggregations, the parameters of edge servers are sent to the cloud for aggregation (with various choices of $E_{\mathsf{Agg}}$ studied).


\textit{3) {\tt \textit{HF-FMs}} with D2D Links ({\tt \textit{HF-FMs + D2D}})}: We follow the same process as the plain {\tt HF-FMs} above; however, one node of each cluster (selected randomly) acts as the cluster head  and operates as follows. After local training, nodes of each cluster send their models via multi-hop D2D links to the cluster head (following the shortest path), which aggregates the received parameters and broadcasts the resulting parameters to the nodes. After $E_{\mathsf{Agg}}$ rounds of local aggregation, the cluster heads upload their parameters to the cloud for aggregation. 

\subsection{Results and Discussion}



Table~\ref{tab:comparison} reports the training latency, energy consumption, and test accuracy of various FFM topologies. It also includes localized variants that perform aggregation \textit{only} at edge servers (without global aggregation), highlighting the role of both localized and global knowledge sharing in model performance.
Compared to the conventional \textit{star topology}, our {\tt HF-FM} and {\tt HF-FM+{D2D}} architectures achieve notable reductions in latency and energy usage, while also yielding higher accuracy due to better personalization at the edge level. In particular, the {\tt HF-FM+{D2D}} variants converge to the same accuracy as their {\tt HF-FM} counterparts (due to replicating the edge aggregations with multi-hop D2D model exchanges) with lower energy and latency, primarily caused by the efficient communication over low-cost, short-range D2D links.
The table also presents the accuracy under varying $E_{\mathsf{Agg}}$ values:  $E_{\mathsf{Agg}}=2$ yields the highest accuracy, highlighting the benefit of balancing rapid edge adaptation with periodic global synchronization. Moreover, less frequent global aggregation (e.g., $E_{\mathsf{Agg}}=8~$ or edge only aggregation) improves latency but comes at the expense of lower  accuracy, underscoring the tradeoff between communication efficiency and model performance.

\setlength{\tabcolsep}{4.4pt} 


\begin{table}[t]
\centering
\caption{Training latency, energy consumption, and test accuracy of our proposed {\tt HF-FM} vs. standard start topology of~FFMs. ~~~~
 $\dag,\triangleleft,\star$ are used to denote the same accuracy performance.}
 \label{tab:comparison}
\begin{minipage}{0.47\textwidth}
\centering
\includegraphics[width=\linewidth]{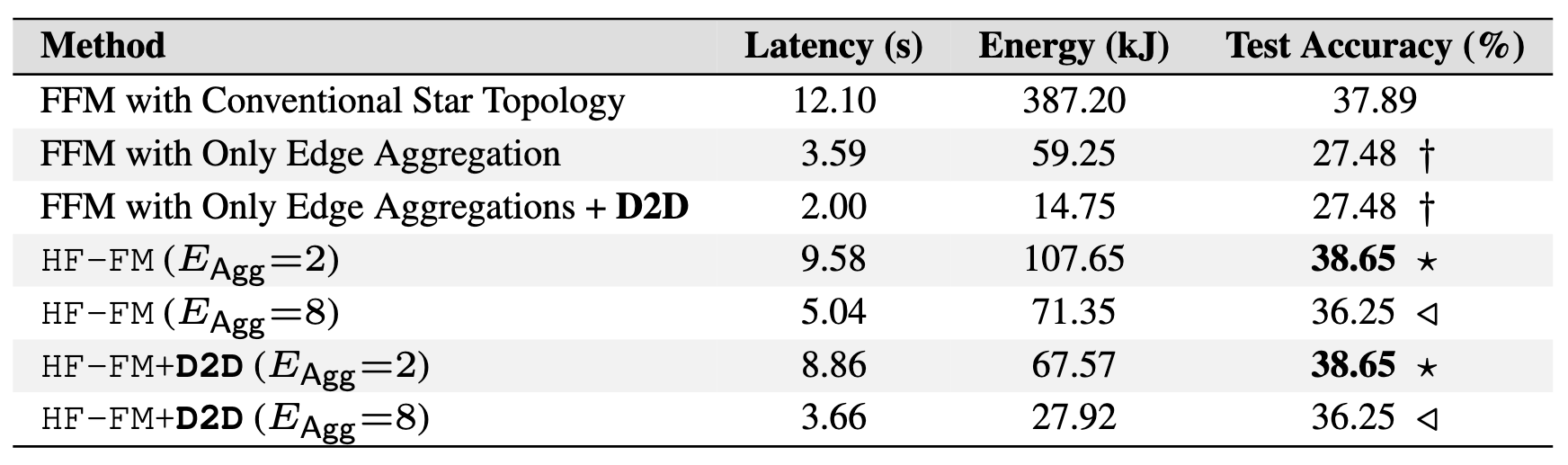} 
\label{fig:yourfig}
\end{minipage}
\hfill
\end{table}

\section{Challenges and Opportunities in {\tt HF-FMs}}\label{sec:challenges}
Our above-described prototype illustrates the potentials of {\tt HF-FMs}. 
Building on this foundation, we now introduce four extended functionalities at both the network and node levels, designed to support principled explorations of this emerging yet impactful research direction (see the color-coded discussions in Fig.~\ref{fig:challenges}).




\subsection{Non-Uniformity of Module Circulation/Traversal} 
Traditional HFL frameworks often apply symmetric aggregation policies, aggregating the entire model or fixed parameter partitions uniformly across nodes (e.g., in split-learning approaches~\cite{thapa2021advancements}).
In contrast, in {\tt HF-FMs}, we introduce non-uniform, module-level aggregations, where each module can be aggregated at various \textit{frequencies/periods} across varying \textit{depth levels} in the fog/edge hierarchy. These aggregation decisions must be tuned according to the module's characteristics (e.g., light-weight prompts vs. parameter-intensive MoEs), nodes' collected modalities, executed tasks, and CCS capabilities. For example, a modality encoder that processes noisy sensor data may benefit from traversing multiple layers (e.g., device-fog/edge-cloud) to reach generalization across disparate sensing environments, especially when higher-tier fog/edge nodes with reliable CCS resources are within reach. Conversely, task heads targeting localized objectives (e.g., urban traffic forecasting) may require shallow aggregation (e.g., at the base station level), preserving region-specific insights.


In addition to such vertical aggregations, {\tt HF-FMs} can exploit \textit{horizontal D2D communications} as a parallel learning pathway: when nearby nodes support short-range connectivity, modules are cooperatively updated through direct D2D exchanges. For instance, a group of nodes can iteratively refine a shared encoder through local consensus, and then forward the refined module to the higher fog/edge layers or relay it to other neighboring nodes for continued adaptation.

\vspace{1mm}
\begin{tcolorbox}[colback=darkyellow2, colframe=teal!40!black, title=\textbf{Future Research Directions}]
\textbf{1. Network-Aware Aggregations:} Developing adaptive algorithms that determine \textit{which modules} to aggregate, \textit{at what frequency and hierarchy depth}, and \textit{via which communication path} (hierarchical, D2D, or hybrid), based on \textit{node attributes}: model personalization needs, CCS resources, data modalities, and executed tasks.

\textbf{2. Model Convergence Analysis:} Establishing theoretical upper/lower bounds to analyze model convergence when its modules are updated asynchronously through hybrid hierarchical--D2D paths, accounting for local D2D topologies, module types, and node attributes.
\end{tcolorbox}

\subsection{Module Relaying for Emerging Task/Modality}
Devices in the {\tt HF-FM} environments often undergo \textit{modality fluctuations} (e.g., when a drone activates or deactivates specific sensors) or \textit{task arrival} (e.g., when a drone joins a surveillance fleet). 
To account for these dynamics, we envision a \textit{module relaying} mechanism in {\tt HF-FMs}, wherein modules are selectively transferred across fog/edge nodes to \textit{accelerate modality/task onboarding and mitigate cold-start overhead} (i.e., eliminating the need to train modules related to new modalities or tasks from scratch).
Module relaying can be carried out both vertically (across hierarchical fog/edge layers) and horizontally (via D2D exchanges), depending on the network topology and resource constraints. 
For instance, a fog/edge node, such as a roadside unit, may hold a high-performing task head module trained on region-specific conditions and proactively relay it to an arriving autonomous vehicle, enabling immediate operation and eliminating the need for redundant local fine-tuning.
Similarly, a node encountering a previously unseen modality (e.g., infrared input) can request and receive a suitable pre-trained encoder from a nearby peer via D2D or from a higher-tier fog/edge node.


\vspace{1mm}
\begin{tcolorbox}[colback=darkyellow2, colframe=teal!40!black, title=\textbf{Future Research Directions}]
\textbf{1. Relay Performance Quantification:} Designing cost functions that assess task/modality similarity and node performance history to determine \textit{which modules} to relay, \textit{from which nodes}, and \textit{through which paths} (hierarchical, D2D, or hybrid), accounting for the CCS, and modality and task heterogeneities of the nodes.


\textbf{2. Module-Level Privacy:} Designing module-level privacy mechanisms that determine \textit{how} to safeguard the exchanged model/data confidentiality against adversarial attacks (e.g., via encryption techniques or injecting calibrated noise to the module parameters).


\end{tcolorbox}

\subsection{Node Specialization}
When a particular module is required for aggregation or relaying, it is impractical to search across a large-scale fog/edge network for all possible combinations of tasks, modalities, node availability, and CCS conditions, as the solution space may scale combinatorially with the network size.
To tackle this, we envision incorporating \textit{node specialization} across the fog/edge network, where instead of having every node handling all modules, certain nodes naturally evolve into \textit{specialists} for specific modules based on the consistency, quality, and relevance of their local data.
For example, a node positioned in a busy urban square may specialize (e.g., via extensive fine-tuning) in modeling pedestrian activity patterns, while another, situated in a quieter residential area, may specialize in analyzing ambient noise levels and air quality fluctuations.
These specialist nodes act as \textit{persistent module providers}, conducting (i) continued fine-tuning and adaptation of their specialized modules using their high-quality data, and (ii) module exchange with other nodes, either directly through D2D links or through the fog hierarchy, when requested.
Notably, node specialization can vary over time as data distributions shift. 
For example, a node aggregating models of wearables might focus on tracking physical activity during the day, and shift to monitoring sleep quality at night.

\vspace{1mm}
\begin{tcolorbox}[colback=darkyellow2, colframe=teal!40!black, title=\textbf{Future Research Directions}]
\textbf{1. Role Governance:}
Designing self-organizing protocols that enable fog/edge nodes to negotiate specialization roles, e.g., based on (i) their local CCS capabilities, along with modality and task coverage,
 and (ii) the historical utility of their modules (e.g., reuse success rates at other nodes).
When multiple nodes converge on the same specialization, developing conflict resolution mechanisms such as distributed voting and performance-based role rotation schemes becomes essential.

\textbf{2. Adversarial Specialists:} Identifying malicious nodes that declare themselves as specialists to gain influence or compensation without contributing meaningful/accurate modules. For example, this can be addressed through cross-validation with overlapping specialists, where multiple specialists claiming the same expertise are compared based on output consistency and accuracy.
\end{tcolorbox}

\subsection{Collaborative Inference}
In the previous subsections, we have assumed that nodes in {\tt HF-FMs} could either \textit{train/fine-tune} specific modules or at least \textit{perform inference} (i.e., feedforward its data through the model to observe the output) using borrowed modules through a locally saved model backbone. In conventional FL/HFL, this assumption often holds due to narrow models that can be stored and executed locally. However, in {\tt HF-FMs}, the sheer size of models introduces a nuanced challenge: some devices may lack the storage or compute power \textit{to even run feedforward inference} locally.
To address this, in {\tt HF-FMs}, we envision \textit{hierarchical, D2D-enabled, collaborative inference approaches} that distribute inference workloads across nearby nodes or escalate them to higher-tier fog/edge layers. For example, when a resource-limited mobile device exists within a trusted user ecosystem (e.g., paired with a personal desktop), inference can be  offloaded to nearby higher-capacity nodes. 



\begin{tcolorbox}[colback=darkyellow2, colframe=teal!40!black, title=\textbf{Future Research Directions}]
\textbf{1. Inference Path Planning:} Developing algorithms to decide \textit{whether} inference can be executed within the user’s ecosystem or needs to be escalated to higher-tier nodes by optimizing the overhead-utility tradeoffs. While vertical escalation through untrusted nodes introduces CCS overhead due to secure communication and cryptographic key exchanges, it can enable the use of higher-quality models at upper layers (e.g., cloud) for improved inference.

\textbf{2. Win-Win Data and Inference Provisioning:} Designing incentivization strategies (e.g., auctions) to unlock the transfer of non-private data of low-capability nodes to more powerful nodes for training/fine-tuning, thereby improving their model qualities.
In return, these powerful nodes can offer inference services back to the contributors under usage contracts, with compensations optimized according to the utility of their shared data.


\end{tcolorbox}

\section{Conclusion}
In this work, we have introduced the paradigm of {\tt HF-FMs}, which synergizes the modular structure of M3T FFMs with the multi-tier architecture of D2D-enabled fog/edge infrastructures. We have developed and open-sourced a prototype implementation of {\tt HF-FMs}, showcasing their promising performance in distributed settings. Finally, by dissecting four key system capabilities, we have outlined a set of open research directions aimed at advancing this untapped area.


\bibliographystyle{IEEEtran}
\bibliography{Ref,New_Bib}
\newpage
\begin{IEEEbiographynophoto}{Payam Abdisarabshali} is a Ph.D. student in EE at SUNY Buffalo, USA.
\end{IEEEbiographynophoto}
\vspace{-180mm}
\begin{IEEEbiographynophoto}{Fardis Nadimi} is a Ph.D. student in EE at SUNY Buffalo, USA.
\end{IEEEbiographynophoto}
\vspace{-180mm}
\begin{IEEEbiographynophoto}{Kasra Borazjani} is a Ph.D. student in EE at SUNY Buffalo, USA.
\end{IEEEbiographynophoto}
\vspace{-180mm}
\begin{IEEEbiographynophoto}{Naji Khosravan} is a senior manager at Adobe Research, USA.
\end{IEEEbiographynophoto}
\vspace{-180mm}
\begin{IEEEbiographynophoto}{Minghui Liwang} is an associated professor of ECE at Tongji University, China.
\end{IEEEbiographynophoto}
\vspace{-180mm}
\begin{IEEEbiographynophoto}{Wei Ni} a Senior Principal Research Scientist and Conjoint Full Professor at Data61, CSIRO,  and the School of CSE, the University of New South Wales, Australia.
\end{IEEEbiographynophoto}
\vspace{-180mm}
\begin{IEEEbiographynophoto}{Dusit Niyato} is a professor of CSE at Nanyang Technological University, Singapore.
\end{IEEEbiographynophoto}
\vspace{-180mm}
\begin{IEEEbiographynophoto}{Michael Langberg} is a professor of EE at SUNY Buffalo, USA.
\end{IEEEbiographynophoto}
\vspace{-180mm}
\begin{IEEEbiographynophoto}{Seyyedali Hosseinalipour} is an assistant professor of EE at SUNY Buffalo, USA.
\end{IEEEbiographynophoto}

\end{document}